\RequirePackage{amsmath}
\documentclass[runningheads]{llncs}
\usepackage[T1]{fontenc}
\usepackage{cite}
\usepackage{orcidlink}
\usepackage{graphicx}
\usepackage{multirow}
\usepackage{subcaption}
\usepackage{xcolor}

\usepackage[capitalise]{cleveref}

\begin{document}
\title{Speech-Guided Sequential Planning for Autonomous Navigation using Large Language Model Meta AI 3 (Llama3)}
\titlerunning{{Speech-Guided Sequential Planning Using Llama3}}
%
\author{Alkesh K. Srivastava\orcidlink{0000-0002-7470-4620}\and
Philip Dames\orcidlink{0000-0002-7257-0075} }
\authorrunning{Alkesh K. Srivastava and Philip Dames}
%
\institute{Temple University, Philadelphia PA 19122, USA\\
\email{\{alkesh,pdames\}@temple.edu}}
\maketitle              
\begin{abstract}
In social robotics, a pivotal focus is enabling robots to engage with humans in a more natural and seamless manner. The emergence of advanced large language models (LLMs) has driven significant advancements in integrating natural language understanding capabilities into social robots. This paper presents a system for speech-guided sequential planning in pick and place tasks, which are found across a range of application areas. The proposed system uses Large Language Model Meta AI (Llama3) to interpret voice commands by extracting essential details through parsing and decoding the commands into sequential actions. These actions are sent to DRL-VO, a learning-based control policy built on the Robot Operating System (ROS) that allows a robot to autonomously navigate through social spaces with static infrastructure and crowds of people. We demonstrate the effectiveness of the system in simulation experiment using Turtlebot 2 in ROS1 and Turtlebot 3 in ROS2. We conduct hardware trials using a Clearpath Robotics Jackal UGV, highlighting its potential for real-world deployment in scenarios requiring flexible and interactive robotic behaviors.

\end{abstract}

\section{Introduction}
\label{sec:introduction}
Social robotics aims to enable robots and humans to cohabitate in a natural and intuitive manner. In the field of collaborative robotics, robots are required to share workspaces with humans, making it essential for robots to understand and execute human commands effectively~\cite{breazeal2016social}. Robots are increasingly being used in retail, industrial settings, households, and office environments. In such settings, robots are often asked to carry out tasks involving multiple steps or to complete a series of tasks in a specific order. For instance, in a retail environment, a robot might be required to navigate the store to pick up items from different aisles and deliver them to a checkout counter. In an office, a robot might need to collect documents from multiple departments and deliver them to a central location, such as Human Resources. At home, robots could be tasked with household chores that involve sequential actions such as serving food from the kitchen and then cleaning up after dinner. Such robotic tasks are often modeled as Vehicle Routing Problems or other optimization challenges~\cite{gonzalez2015review}. However, the complexity of these problems increases in dynamic environments, with more assigned tasks, and a growing number of robots. Robots are also increasingly deployed for these tasks in hostile or hard-to-reach areas~\cite{bellingham2007robotics}. Solving these issues is challenging on its own, and incorporating natural language understanding for collaborative robotics adds an additional layer of difficulty.

This paper aims to enable people to sequentially task robots using natural language inputs. However, in experimental robotics, models are often designed for specific tasks with rigid input formats to ensure successful task execution. This rigidity ensures that robots can accomplish their assignments accurately but does not account for the natural variability in human communication. Humans, celebrating their individuality, interact with robots in diverse and natural ways, often not conforming to predefined input formats. Therefore, for collaborative robotics to be truly successful, robots must be able to interpret a variety of human commands and translate them into actionable plans~\cite{ren2023robots}. To address this need, we will build upon recent advancements in large language models, which have shown great promise in enhancing natural language understanding. 
\begin{figure}[!ht]
    \centering
    \includegraphics[width=.9\textwidth]{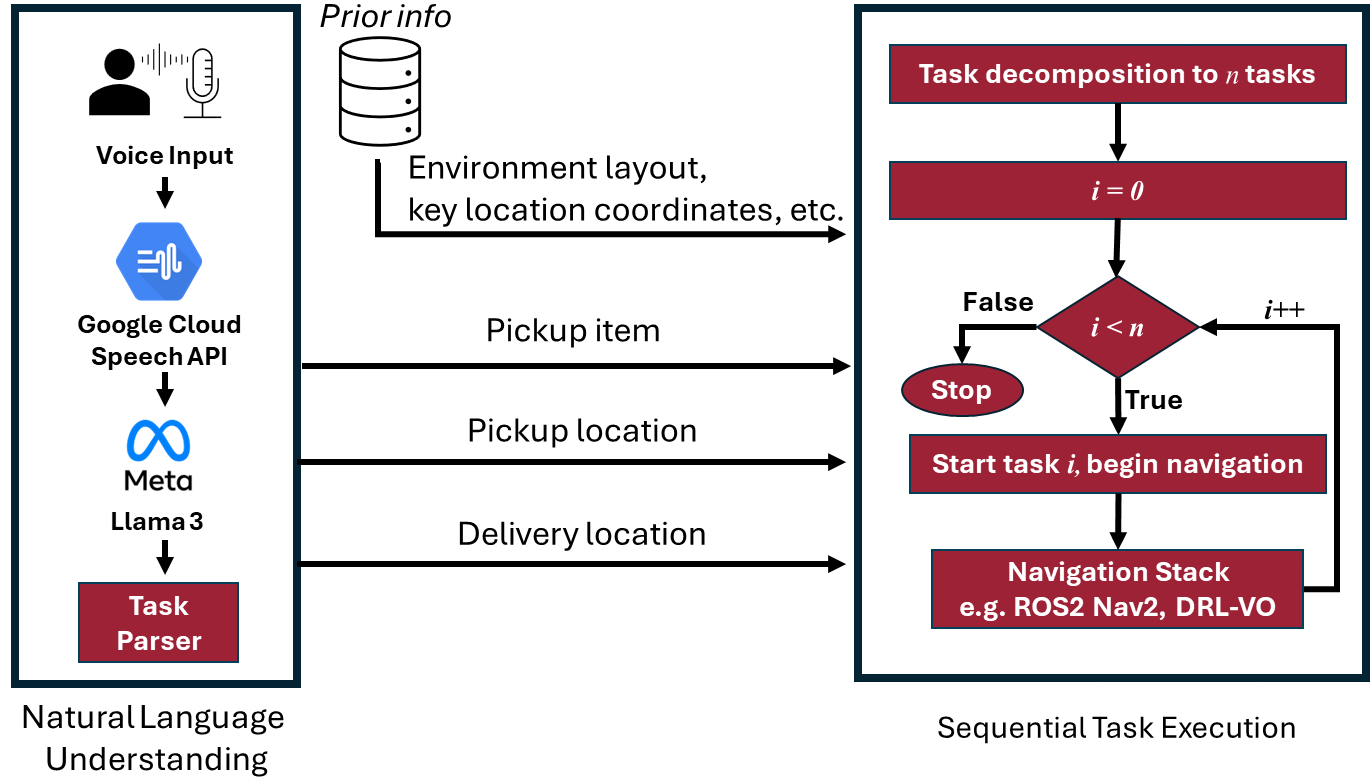}
    \caption{Overview of the proposed system. The process begins with converting verbal commands into text using the Google Cloud Speech API. The Llama3 model processes this text to extract essential details, such as the pickup location, item, and delivery destination, using regex-based parsing. These parsed commands, along with prior environmental information, are sent to the Task Execution module, where they are translated into a sequence of actions that are then executed.}
    \label{fig:overview}
\end{figure}

Our system, illustrated in \cref{fig:overview}, consists of two main modules. The first is the natural language understanding module, which processes user speech or text input and outputs a set of task parameters. We utilize speech-to-text technology to convert verbal commands into text. The text is fed into Llama3~\cite{llama3}, an autoregressive large language model, which extracts essential details—such as pickup location, item, and delivery destination—using regex-based parsing. In other words, we are using Llama3 to ``translate'' free-form natural language commands into a standardized, easily parsable form for robust task extraction.
Although alternatives like BERT~\cite{devlin2018bert} could be trained on large datasets to handle this process, they would likely eliminate the need for an LLM. Instead, we prioritize Llama3’s ability to generalize and flexibly interpret commands, which allows us to maintain a simpler, rule-based parsing mechanism while still benefiting from the LLM’s advanced linguistic capabilities.

The second module translates the parsed commands into a sequence of actions by referencing a predefined dictionary of environment coordinates. This dictionary stores coordinates of key landmarks, eliminating the need for semantic planning and enabling faster execution. Task execution is managed using a finite-state machine to control the flow of operations, along with off-the-shelf robot navigation algorithms.

We demonstrate the efficacy of our system through simulated and hardware experiments using three different robot models, two navigation algorithms, and two environments. These experiments highlight the modularity of our system and the potential for practical applications in real-world scenarios.



\section{Related Work}
\label{sec:related_work}
In this section, we review the literature related to each of our two modules: natural language understanding (NLU) and robotic navigation.

\subsection{Natural Language Understanding}
Recent advancements in large language models (LLMs) have significantly enhanced natural language understanding (NLU) capabilities across various domains, including robotics~\cite{zhao2023survey}. Models like GPT-3 and BERT~\cite{devlin2018bert} have demonstrated exceptional performance in comprehending and generating human-like text, owing to their extensive training on large datasets. By fine-tuning these models for specific tasks, researchers have achieved state-of-the-art results in tasks spanning from text classification to question answering~\cite{zhao2023survey, zhou2023comprehensive}, thereby enhancing their performance on conventional NLU benchmarks~\cite{chang2024survey, huang2022towards}. SayPlan~\cite{rana2023sayplan} focuses on scalable task planning using 3D scene graphs but does not integrate speech-recognition or incorporate socially compliant algorithms such as DRL-VO~\cite{xie2023drlvo} for plan execution.

LLMs excel particularly in tasks requiring broad contextual understanding or dealing with unstructured data~\cite{chang2024survey}, which is pertinent to collaborative and social robotics applications. LLMs exhibit generalization capabilities, surpassing traditionally fine-tuned models in handling diverse and adversarial inputs~\cite{liang2022holistic, nie2019adversarial}. In terms of practical applications, recent studies have showcased LLM effectiveness across various NLU tasks, such as machine translation~\cite{brants2007large}, question answering~\cite{singhal2023towards}, and text classification~\cite{chae2023large}.

\textbf{Contributions} \:
The novelty of our framework lies in its integration of speech recognition and NLU capabilities, leveraging LLMs to enhance task classification accuracy. Our approach seamlessly integrates Google Speech Cloud API for speech-to-text transcription with Llama3 for task classification.

\subsection{NLU-guided Social Autonomous Navigation}
The vision of seamless integration of mobile robots into human environments has been extensively studied. Robots like RHINO and MINERVA were deployed in museums and solely focused on autonomous navigation amidst humans~\cite{lakemeyer1998interactive}. Various approach of navigation has treated people as dynamic, non-responsive obstacles, and emphasized collision avoidance~\cite{vasquez2014inverse}, while other approaches integrated human motion prediction with robot decision-making, acknowledging mutual influences between actions of robots and people~\cite{fridovich2020efficient}.

Recent advancements in robotic navigation guided by natural language have explored diverse methodologies to enhance human-robot interaction and adaptability in dynamic environments. FollowNet~\cite{shah2018follownet}and LM-Nav~\cite{shah2023lm} leverage end-to-end neural architectures and large pre-trained models to interpret natural language instructions and navigate complex environments, demonstrating success in simulated and real-world scenarios. GOAT~\cite{chang2023goat} offers a multimodal navigation system that integrates language descriptions and object recognition. However, it lacks a comprehensive approach to dynamic social interactions in complex environments. Arena 3.0~\cite{kastner2024arena} provides a realistic simulation environment for social navigation but does not incorporate the advanced NLU features and real-time feedback mechanisms that the proposed approach offers.

\textbf{Contributions} \:
Our work advances beyond existing methodologies by combining robust NLU capabilities with adaptive navigation strategies. 
Unlike previous approaches, which often rely on fixed control policies or simplistic command parsing techniques, our methodology leverages the capabilities of Llama3 for robust natural language understanding and DRL-VO \cite{xie2023drlvo} for adaptive and socially-compliant navigation.

\section{Problem Statement}
\label{sec:problem_statement}
The general task addressed in this work is to understand human commands using Large Language Models (LLMs) and form a sequential plan for execution. The overall problem can be subdivided into two problems:

\subsection*{Problem 1: NLU for Command Interpretation}
This paper focuses on the scenario where a social robot receives instructions to pick up an object from a specified location and deliver it to another designated place. This task exemplifies a common class of multi-step operations in social robotics, where the robot must navigate through sequential actions such as in household assistance for cleaning tasks, delivering medication in health care settings, etc. By addressing pickup and delivery planning, we aim to tackle fundamental challenges that underpin various practical applications.

Let ${C}$ denote the natural language command (in text or speech) provided by the user, which includes information about pickup location $L_{\rm pickup}$, delivery location $L_{\rm delivery}$, and the pickup item $I$.  The task of the robot is to design a Natural Language Understanding (NLU) system that accurately parses $C$ into a structured format:
\begin{equation*}
    L_{\rm pickup}, L_{\rm delivery}, I = \textsc{nlu}({C}),
\end{equation*}
where $\textsc{nlu}(C)$ process $C$ using {Llama3} and parses the obtained information to extract the task parameters $L_{\rm pickup}, L_{\rm delivery},$ and $I$.

\subsection*{Problem 2: Autonomous Navigation } 
We assume an environment whose layout of the static infrastructure is known and we possess coordinates for key locations within it (e.g., room numbers). The robot shares the environment with people who may move around during the robot's operation.
The robot model used in this study is a differential wheel drive robot, which is a representative model of various ground robots such as TurtleBots, Jackal UGV, and Moxi by Diligent Robotics, though the proposed system is applicable to any mobile robot. 

Once the task parameters are identified by the \textsc{NLU}, we can look up the coordinates of the pickup and delivery locations, $L_{\rm pickup}$, $L_{\rm delivery}$, in the environment map. The next challenge is to navigate through the environment to reach those locations. The robot does this using a control policy $\pi$ with parameters $\theta$ that selects steering actions $a_t$ based on partial environmental observation $o_t$ (obtained from the sensors and perception system):
\begin{equation*}
    a_t \sim \pi_\theta(a_t|o_t).
\end{equation*}

\section{Methodology}
\label{sec:methodology}
In this section, we describe the methodology of the proposed Speech-Guided Sequential Planner for Autonomous Navigation. The proposed system integrates speech recognition, natural language understanding, and an advanced control technique to enable a robot to autonomously navigate in an environment populated with human pedestrians. 
First, we address the natural language understanding aspect of Problem 1 in \cref{subsection:nlu}. Followed by addressing the autonomous navigation challenge posed by Problem 2 in \cref{subsection:tasks}.

\subsection{Natural Language Understanding (NLU)}
\label{subsection:nlu}
The first module takes in a natural language input (speech or text) and parses it to extract the task information. This is done in two steps.

\subsubsection{Speech To Text Conversion}
\label{subsection:speech2text}

To convert speech to text, we utilize the Google Speech Recognition API~\cite{google_speech_recognition} through the Python3 \texttt{speech\_recognition} library for precise speech-to-text conversion. Initially, ambient noise levels are calibrated using the microphone's input during the first second without processing speech, adjusting the recognizer's sensitivity accordingly. Google's advanced speech recognition technology employs signal processing and machine learning methods. It begins by preprocessing the audio input to extract spectral representations and phonetic patterns. Deep neural networks then map these features to textual sequences with high accuracy and efficiency.

\subsubsection{Large Language Model (LLM)}
\label{subsection:llm}

The natural language understanding (NLU) process encompasses several key steps. Firstly, the received text undergoes preprocessing, which includes tasks such as punctuation removal and text formatting to maintain consistency throughout the text. 
Following preprocessing, the text is fed into the Llama3 API~\cite{llama3} provided by Groq~\cite{groq} for context understanding and semantic parsing.\footnote{We employ the 8 billion parameter version (\texttt{llama3-8b-8192m}) instead of the larger counterparts, such as the 80 billion parameter model, due to larger models exhibiting a tendency to assign specific room numbers when interpreting generic commands. For instance, if asked to navigate to ``TRAIL lab,'' it might erroneously label it with a hallucinated room number, like ``Room 111.'' Additionally, we can use the smaller model onboard the robot rather than relying on remote API services, leading to quicker response times and greater practicality for real-time applications.} 
This step involves extracting structured data essential for task execution. The extracted information is parsed using regular expressions to identify the pickup location \( L_{\rm pickup} \), delivery location \( L_{\rm delivery} \), and the item to be picked up \( I \). We utilize the LLM as a translation layer to reformat complex, natural language commands into a standardized, structured format that simplifies parsing. The integration of Llama3 enhances the system's capability to accurately interpret and process natural language commands, promoting seamless interaction and task execution in robotic applications. Opting for regex-based parsing over learned task classifiers ensures greater flexibility in handling diverse human commands and reduces computational costs, making it more practical for implementation in robots.

\subsection{Sequential Task Assignment and Execution}
\label{subsection:tasks}
Once the command is interpreted, we formulate and execute the task sequence.

\subsubsection{Task Assignment}
For task assignment, our system employs a Finite State Machine (FSM) for handling simple tasks such as a single pickup and delivery. The FSM consists of states like Idle, Navigating to Pickup, Picking Up Item, Navigating to Delivery, and Delivering Item, with transitions triggered by events such as reaching a location or completing an action. This approach is efficient for straightforward tasks. However, for more complex tasks that involve multiple steps and dependencies, we can employ hierarchical task planners like GTPyhop~\cite{nau2021gtpyhop} or SHOP~\cite{nau2001shop}. These planners decompose high-level tasks into manageable sub-tasks, enabling the robot to handle complex scenarios, thus justifying their integration into our system for enhanced task management and execution.

\subsubsection{Autonomous Navigation}
\label{subsection:drlvo}
We utilize two different navigation frameworks, depending on the situation. In static worlds (i.e., without people), we use the ROS2-Nav2 navigation stack. This is one of the most commonly used navigation frameworks in mobile robots.
In environments full of people, we utilize the DRL-VO (Deep Reinforcement Learning with Velocity Obstacles)~\cite{xie2023drlvo} navigation system, which the authors previously designed to enable robots to navigate through crowded and dynamic environments. The control policy is a convolutional neural network that uses lidar scans, maps of pedestrian locations/speeds, and goal coordinates to generates velocity commands for the robot. DRL-VO yields higher speeds and few collisions than other robot controllers (including the ROS navigation stack), especially in dense crowds.

\section{Experiments}
\label{sec:experiments}

\begin{figure}[tbph]
    \centering
    \includegraphics[width=1\textwidth]{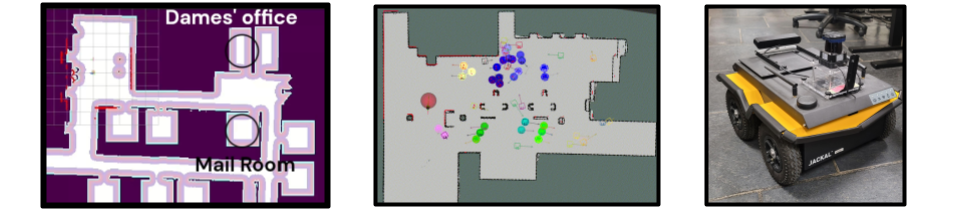}
    \caption{(Left) The layout of the Mechanical Engineering department at Temple University, used in the ROS2-Nav2 simulation with Turtlebot3. (Center) The layout of the lobby of the College of Engineering at Temple University, which is used for both simulation experiments with Turtlebot2 using DRL-VO and hardware experiments with the Jackal UGV. (Right) The Jackal UGV.}
    \label{fig:experimental_setup}
\end{figure}

To evaluate the system, we first assess the speech-to-task classification by having volunteers self-report the accuracy of the classification on 10 natural task statements. We then evaluate the system integration with three distinct robotic experiments: a simulation in an empty office environment using a Turtlebot3 in ROS2, a simulation in a lobby with people using a Turtlebot2 in ROS1, and a hardware implementation in a lobby with people using a Clearpath Robotics Jackal UGV. The experimental setup is described in~\cref{fig:experimental_setup}. These experiments collectively demonstrate the system's versatility and effectiveness across different scenarios, from pedestrian-free areas to crowded social spaces.

\subsection{Speech-to-Task Accuracy}
To test the efficacy of the natural language understanding aspect of our system, we asked 10 volunteers to speak 10 commands and report the number of correct classifications. Out of the 10 commands, 5 were provided to them and the rest were the volunteer's own command statements. This ensures that the accuracy reported by the system is unbiased. Examples of the commands include ``Could you please bring the keys from security to TRAIL?'' and ``I forgot my laptop, please bring a laptop from the computer station to the robotics lab.'' We observed an average accuracy of $84.37\%$ for task classification. Feedback from the volunteers revealed that using identifiers like ``the'' or ``a'' before the location or pickup item confused the NLU system, suggesting a need for improved preprocessing to handle such variations in command phrasing.

\begin{figure}[tbph]
    \centering
    \begin{subfigure}[b]{0.19\textwidth}
        \centering
        \includegraphics[width=\textwidth]{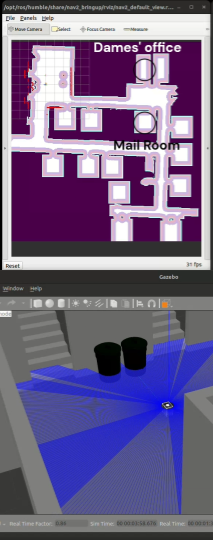}
        \caption{Step 1: Localizing robot's current location in the environment (upper left room).}
        \label{fig:subfig1}
    \end{subfigure}
    \begin{subfigure}[b]{0.19\textwidth}
        \centering
        \includegraphics[width=\textwidth]{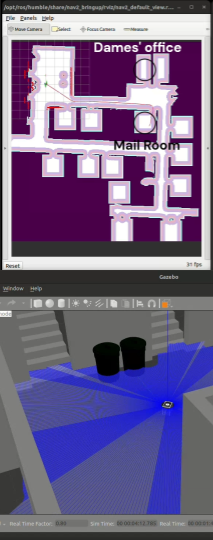}
        \caption{Step 2: Planning path and navigating to the $L_{\rm pickup}$ (Mail Room).}
        \label{fig:subfig2}
    \end{subfigure}
    \begin{subfigure}[b]{0.19\textwidth}
        \centering
        \includegraphics[width=\textwidth]{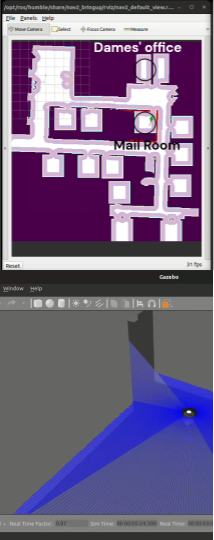}
        \caption{Step 3: Reach $L_{\rm pickup}$ and executing action for pickup $I$ (envelopes).}
        \label{fig:subfig3}
    \end{subfigure}
    \begin{subfigure}[b]{0.19\textwidth}
        \centering
        \includegraphics[width=\textwidth]{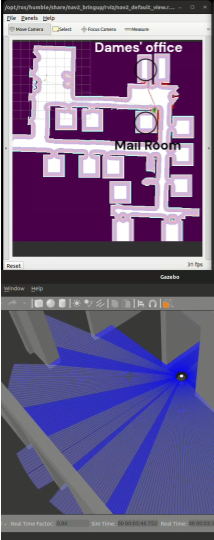}
        \caption{Step 4: Planning path and navigating to the $L_{\rm delivery}$ (Dames' Office).}
        \label{fig:subfig4}
    \end{subfigure}
    \begin{subfigure}[b]{0.19\textwidth}
        \centering
        \includegraphics[width=\textwidth]{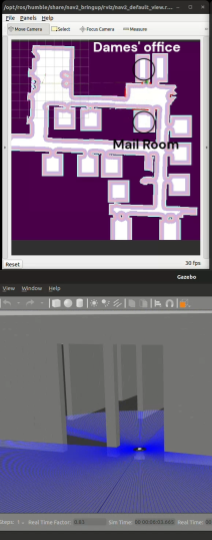}
        \caption{Step 5: Reach $L_{\rm delivery}$ and deliver $I$ to complete the sequential tasks.}
        \label{fig:subfig5}
    \end{subfigure}
    \caption{Simulation experiment depicting the steps of progress for pickup of item $I$ (envelopes) from $L_{\rm pickup}$ (Mail Room) and its delivery to $L_{\rm delivery}$ (Dames' Office).}
    \label{fig:main}
\end{figure}

\subsection{System Integration Testing}

We conducted three tests, outlined in \cref{tab:experiments} that vary the robot platform, environment, and ROS version to demonstrate the flexibility of the system. The table also provides links to full videos of all experiments, and \cref{fig:main} shows a series of snapshots of the first experiment with the Turtlebot 3. The videos show that in all cases the robot is able to complete the task given to it, either from a natural language text input on a remote computer or a spoken command. The simulated and hardware tests in the lobby environment both have pedestrians, demonstrating the system's capability to navigate autonomously in social spaces with static infrastructure and moving crowds, which is essential for getting robots outside of controlled lab and factory environments. 

\begin{table}[tbph]
    \centering
    \caption{Overview of Experiments}
    \label{tab:experiments}
    \smallskip
    \begin{tabular}{c|c|c|c|c}
        Type & Robot & Navigation & Environment & Videos \\ \hline \hline
        \multirow{2}{*}{Simulation} & \multirow{2}{*}{Turtlebot3} & \multirow{2}{*}{ROS2-Nav2} & \multirow{2}{*}{ME Dept.} & \url{https://youtu.be/L3kIdW80ZK0} \\
        & & & & \url{https://youtu.be/HGL9E_DZUsk} \\ \hline
        \multirow{2}{*}{Simulation} & \multirow{2}{*}{Turtlebot2} & \multirow{2}{*}{DRL-VO} & \multirow{2}{*}{Lobby} & \url{https://youtu.be/TfleIxcCoE8}\\
        & & & & \url{https://youtu.be/BaoLZ68bkAM} \\ \hline
        Hardware & Jackal & DRL-VO & Lobby & \url{https://youtu.be/hzYRvpX9Qe8}
    \end{tabular}
\end{table}

\subsection{Discussion}
\label{sec:discussions}
Our experimental setup aimed to demonstrate the versatility and compatibility of our system with different robotic platforms and environments. In a controlled, pedestrian-free environment, we utilized the ROS2 Nav2 package with Turtlebot3, emphasizing the system's capability to operate effectively with smaller base robots and showcasing its integration with advanced ROS2 features. In human-populated environments, where the risk of collision is higher, we used the DRL-VO \cite{xie2023drlvo} algorithm on a Turtlebot2 and a Jackal. This highlights our system's adaptability to various robotic platforms, software stacks, and environments to be able to operate in a range of settings.

\section{Conclusion}
\label{sec:conclusion}
In this paper, we explore the feasibility and effectiveness of integrating advanced natural language processing with robotic navigation systems to enable people to use natural language to ask robots to complete multi-step tasks. We leverage Llama3 to parse commands and different off-the-shelf navigation algorithms, and we apply our system to a range of robot models, software stacks, and environments. Our research enhances social and collaborative robotics by enabling people to naturally interact with robots in shared spaces. 

Future work will aim to enhance the versatility to more real-world scenarios.
We will update the robotic hardware by adding a gripper to the robot. This will allow it to pick up and deliver objects as well as autonomously utilize elevators, presenting an intriguing opportunity to expand their operational capabilities in multi-story buildings and complex indoor environments.
We will also update the software with the goals of: 1) enhancing the success rate of our NLU module, 2) allowing the robot to respond to the user if it does not understand the given query \cite{ren2023robots}, and 3) broadening the scope of tasks, which may include additional steps to complete, by utilizing hierarchical task networks (HTNs) \cite{nejati2006learning} or similar planning paradigms.

\section*{Acknowledgment}
This work was funded by NSF grant CNS-2143312. The authors would like to thank Zhanteng Xie, Jared Levin, and Alexander Derrico for their assistance with the hardware experiments.

%
%
%
\bibliographystyle{splncs04}
\bibliography{mybibliography}

\end{document}